\documentclass[11pt,a4paper]{article}

\usepackage{cloudrobo}
\usepackage{float}          % 新增：用于 [H] 固定图片位置
\renewcommand{\CloudRoboName}{CloudRobo}

\renewcommand{\CloudRoboWordmark}{\CloudRoboName}

\definecolor{CloudRoboAccent}{HTML}{512FD6}
\definecolor{CloudRoboInk}{HTML}{512FD6}
\definecolor{CloudRoboMuted}{HTML}{665F77}
\definecolor{CloudRoboRule}{HTML}{C9BDEF}
\definecolor{CloudRoboTint}{HTML}{F4F1FD}

\renewcommand{\CloudRoboWordmarkFont}{\sffamily\bfseries\fontsize{19}{22}\selectfont}

\CloudRoboRepeatLogotrue

\renewcommand{\CloudRoboTitleStyle}{classic}
\definecolor{CloudRoboTitleColor}{HTML}{000000}

\title{
CR-VLA-Force: Learning Control-aware Compliance VLA Model for Robust Contact-rich Robotic Manipulation}

\author{
Zhaohong Mai\textsuperscript{2}, 
Chao Wang\textsuperscript{1,$\dagger$}, 
Chao Zeng\textsuperscript{2}, 
Sitong Mao\textsuperscript{1}, 
Heng Zhang\textsuperscript{1,*,$\dagger$}\\
Shunbo Zhou\textsuperscript{3}, 
Chenguang Yang\textsuperscript{4,*,$\dagger$}
\\[0.3em]
{\normalfont\small
\textsuperscript{$\dagger$} Project Leader \qquad
\textsuperscript{*} Corresponding Author (zhangheng203@huawei.com, cyang@ieee.org)}\\
{\normalfont\small
\textsuperscript{1} Huawei Technologies Co., Ltd.
\qquad\\\textsuperscript{2} South China University of Technology
\qquad
\textsuperscript{3} Ola Dimensions
\qquad
\textsuperscript{4} The Hong Kong Polytechnic University
}
}

\date{}

\renewcommand{\CloudRoboRunningTitle}{CloudRobo Technical Report}

\renewcommand{\CloudRoboAffiliation}{}

\renewcommand{\CloudRoboLinks}{%
  \href{https://www.huaweicloud.com/product/cloudrobo.html}
  {Project page}
}

\renewcommand{\CloudRoboKeywords}{
control-aware, force/torque, VLA models, compliance control, contact-rich
}

\begin{document}

\maketitle

% ============================================================
% Framework figure: placed between title and abstract
% ============================================================
\begin{figure}[H]
  \centering
  \includegraphics[width=1.0\textwidth]{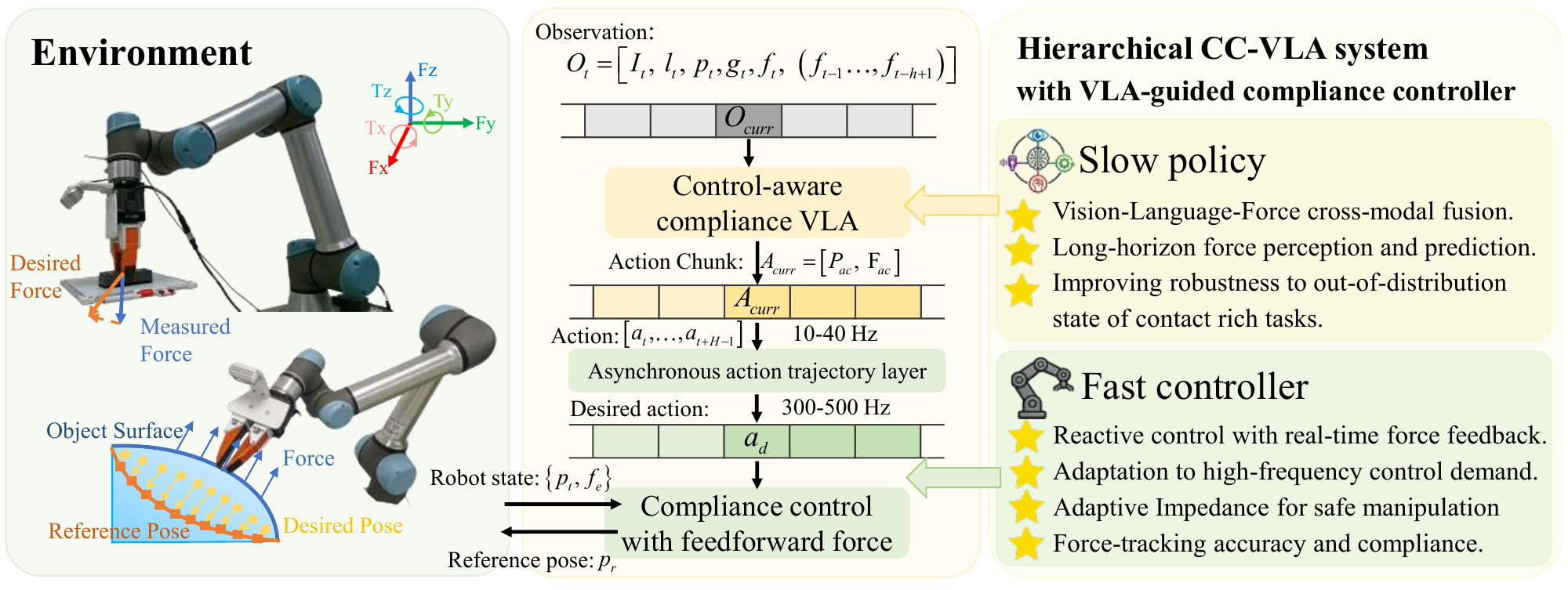}
  \caption{
    The CC-VLA framework overview: the VLA model
    (slow policy) integrates vision, language, force sequences and
    proprioceptive states to infer action chunks. Meanwhile, the
    VLA-guided compliance controller (VG-ACC) ensures effective
    tracking of both the feedforward force and the desired pose.
  }
  \label{fig:architecture}
\end{figure}

% ============================================================
% Abstract
% ============================================================
\begin{abstract}
Integrating visuomotor policies or Vision-Language-Action (VLA) models with force/torque (F/T) perception has demonstrated significant progress in imitation learning for robotic manipulation. However, existing force-aware VLA models frequently exhibit limited capability in precise force tracking and rapid successive adjustments. This deficiency stems from the limitations of action-chunk execution strategies and the substantial latency between perception and real-time control. Such limitations can lead to task failures and safety risks, particularly when the execution of an action chunk exerts excessive interaction forces without timely adjustment. To overcome this challenge, we propose the Control-aware Compliance VLA (CC-VLA) framework for reactive control. The CC-VLA model employs a multimodal mixture-of-experts (MoE) to encode force signal sequences and vision-language fused feature. Furthermore, it utilizes a multi-stage training strategy to ensure robust perception within the visual-semantic space and effective force perception under sparse sampling conditions. Additionally, a VLA-guided adaptive compliance controller is designed to facilitate precise position tracking during contact-free motion and optimal force-position tracking for contact-rich tasks. To facilitate high-precision F/T data acquisition, we also implement an adversaria shared teleoperation strategy for contact-rich demonstrations that bolsters system safety and interactivity. Extensive real-world experiments demonstrate that CC-VLA significantly improves success rates in challenging force-perception tasks and enhances force-control precision, while providing multi-level safety and robustness under the tested partial-OOD pose-shift settings.

\end{abstract}

% ============================================================
% Main body
% ============================================================
\section{Introduction}
\label{sec:introduction}
Contact-rich robotic manipulation spans many practical tasks but remains challenging \cite{suomalainen2022survey}. In out-of-distribution (OOD) settings, dynamic scenes, or under external disturbances, minor perception or control errors can trigger force transients, unintended collisions, or contact loss, leading to unsafe robot–environment interactions. Inspired by human-like learning controller \cite{yang2011human} and passivity-enforcement techniques \cite{landi2017admittance}, research on variable compliance control has gained substantial momentum as a means of ensuring stability \cite{kronander2016stability}.

With advancements in generative imitation learning utilizing action chunking, visuomotor policies (e.g., ACT \cite{zhao2023learning} and diffusion policy (DP) \cite{chi2025diffusion}) and VLA models (e.g., ${\pi _0}$ \cite{black2410pi0} and ${\pi _{05}}$ \cite{intelligence2025pi_}) have established a strong methodological basis for end-to-end robotic manipulation. However, without adequately incorporating control knowledge, such as control feasibility, compliance mechanisms, and stability constraints, these end-to-end policies typically exhibit limited stability and robustness in contact-rich tasks. Although ForceVLA \cite{yu2025forcevla} and TA-VLA \cite{zhang2025elucidating} incorporate F/T modalities into VLA, their applicability to contact-rich interactions remains limited by the control frequency typical of VLA models. In particular, they primarily employ position-based control without explicit or implicit force control, making it difficult to meet the accuracy requirements. Consider a plug insertion task: the robot must apply sufficient force to fully seat the plug while remaining compliant and able to rapidly adjust its motion; otherwise, a slight misalignment can induce excessive contact forces and cause damage. To address these challenges, incorporating compliance control priors into VLA models seamlessly integrates the adaptability and generalization of learning systems with the reliability and precision of safety-aware controllers.

It can be observed that vision and force are highly heterogeneous in terms of semantic abstraction, information density, sparsity, and temporal dynamics \cite{he2025foar}. Direct end-to-end fusion often leads to modality dominance, whereby the model over-relies on the information-rich visual modality and underutilizes force cues. Furthermore, introducing force signals too early is suboptimal, as their inherent noise can degrade the learning of generic spatial perception. From a control-aware perspective \cite{zeng2025learning}, force is better regarded as both proprioceptive input and an action target \cite{zhang2025elucidating}, serving as a strong constraint or corrective signal. Prior work, including ControlNet \cite{zhang2023adding} in image generation, AV-HuBERT \cite{shi2022learning} in audio–visual speech recognition, and CT–MRI fusion \cite{dou2020unpaired} in medical imaging, demonstrates that multi-stage training with auxiliary branches is effective for multimodal integration under conditions of imbalance. Inspired by Forcevla \cite{yu2025forcevla}, we adopt a multi-stage strategy for MoE-based \cite{mustafa2022multimodal} vision–semantics and force fusion: Stage I trains the VLA backbone for robust perception and planning, while Stage II trains the MoE to exploit force residuals for action correction. To construct a physically grounded, controller-aware latent space for the VLA and to meet the requirements of the controller, both F/T and pose are used as prediction targets for the action expert.

Contemporary VLA models typically predict actions from a single observation frame, implicitly assuming a Markovian environment. However, a single frame is often insufficient to infer task progress or completion in non-Markovian tasks, which can result in missed steps or erroneous actions. To mitigate this limitation, MemoryVLA \cite{shi2025memoryvla} maintains a retrievable memory that integrates low-level perceptual details with high-level semantic progress. Likewise, the Point Tracking History-Aware Policy \cite{chen2025history} compresses temporal context into point trajectories of task-relevant objects and enables asynchronous tracking via cross-attention. The importance of history is particularly evident in contact-rich manipulation: sequences of F/T measurements capture both magnitudes and temporal variations, providing contact-state cues that are unavailable from images or instantaneous F/T readings alone. Motivated by these observations, we propose a force-aware causal attention module that encodes F/T sequences into learnable embeddings and supplies them as inputs to a cross-modal fusion expert.

With the above discussion, we highlight the contributions of this paper as follows.

\begin{enumerate}

  \item The proposed control-aware compliance VLA unifies force awareness and compliant force control within a hierarchical framework. Specifically, it employs a cross-modal fusion expert and a multi-stage training strategy to integrate force and vision-language modalities, enabling action prediction that explicitly incorporates force feedforward for compliance control.

  \item We propose a historical force sequence encoder integrated with a real-time force feature mapper to constitute the feature fusion network.

  \item Addressing the demands of high-frequency control, safety, and compliant interaction in contact-rich tasks, we propose a VLA-guided adaptive compliance controller that utilizes the desired positions and feedforward forces predicted by the control-aware VLA.

\end{enumerate}

Finally, we evaluate the efficacy of the proposed CC-VLA framework across a spectrum of demanding contact-rich tasks. Experimental results indicate that our approach not only achieves a marked increase in success rates and force-position tracking precision but also maintains high performance in partial out-of-distribution scenarios, highlighting its superior robustness. The CC-VLA framework is illustrated in Fig. 1.

\section{RELATED WORK}

\subsection{End-to-End force/torque aware policy learning}

Recent work has begun to learn compliance and force regulation end‑to‑end within visuomotor and VLA policies for real‑world manipulation (e.g., wiping and insertion). Adaptive Compliance Policy \cite{hou2025adaptive} and its UMI‑FT \cite{choi2026wild} extension employ a diffusion policy conditioned on images, pose, and F/T signals to predict anisotropic stiffness and a reference pose, which are executed via a variable compliance controller. This differs from the adaptive stiffness computation used in this work. Other diffusion‑based methods (e.g., ForceMimic \cite{liu2025forcemimic} and FoAR \cite{he2025foar}) similarly fuse vision, proprioception, and F/T measurements to jointly predict future motion. TacDiffusion \cite{wu2025tacdiffusion} and FILIC \cite{ge2025filic} further close the loop through compliance control. To address the frequency mismatch between slow visual feedback and fast contact sensing, Reactive Diffusion Policy \cite{xue2025reactive} adopts a slow–fast hierarchical design, whereas ImplicitRDP \cite{chen2025implicitrdp} integrates both timescales into a diffusion model via slow–fast causal attention and virtual‑target regularization. Simulation‑based external force estimation and policy learning can also be integrated with hybrid force–position control during sim‑to‑real deployment \cite{zhi2025learning}.

However, integrating F/T information into VLA models remains an open challenge. Prior visuomotor policies typically use self-attention \cite{choi2026wild} or cross-attention \cite{ge2025filic} to fuse visual and F/T signals. TA-VLA \cite{zhang2025elucidating} treats force as part of the proprioceptive state and incorporates the desired force prediction error into the training objective. ForceVLA \cite{yu2025forcevla} uses a MoE block to fuse vision–language representations with real-time force signals, thereby capturing phase-dependent interaction dynamics and conditioning the action head of \cite{black2410pi0} to generate contact-aware robot actions. Although these VLA models can generate pose action chunks conditioned on interaction forces, they remain limited in achieving accurate desired force tracking and reactive force–position adjustments under sparsely sampled observations.

\subsection{Compliance control with impedance learning}
Inspired by human motor learning \cite{yang2011human}, variable impedance control (VIC) is fundamental to robust contact-rich manipulation. Learning from demonstration is widely used to encode desired actions such as trajectories and force profiles \cite{zhao2022hybrid,zeng2024hierarchical}. To reproduce these skills, optimization-based strategies (e.g., quadratic programming (QP)) are frequently employed to adapt stiffness online \cite{zeng2024hierarchical}, often incorporating energy-tank constraints to ensure passivity and stability \cite{zhao2022hybrid,beber2024passive}.

Concurrently, data driven approaches have advanced to improve generalization and safety. Inverse reinforcement learning has been applied to recover impedance policies and reward functions, demonstrating superior transferability compared with behavior cloning \cite{zhang2021learning}. Likewise, safe reinforcement learning frameworks integrate VIC to modulate stiffness dynamically, thereby enabling safe exploration in contact-rich tasks \cite{zhang2024srl}. Continuous learning and biomimetic control enable robots to autonomously explore and acquire new skills \cite{zeng2025robot}. Most recently, diffusion-based models have been proposed to reconstruct contact-consistent, zero-force trajectories, effectively bridging information-driven motion generation with energy-based impedance regulation \cite{geiger2025diffusion}. Despite significant advances, the design of variable compliance controllers that are closely coupled with control-aware multimodal policies remains an outstanding research challenge.

\section{METHODOLOGY}

The proposed control-aware compliance VLA is designed as a slow–fast hierarchical system composed of a multimodal robotic policy and an adaptive compliance controller, as illustrated in Fig. 2. Built upon the ${\pi _0}$ foundation model \cite{black2410pi0}, it employs PaliGemma to encode visual observations and task instructions into contextual embeddings, and uses a flow-based action expert network to predict action chunks.

To effectively integrate force-related signals with visual semantic information, the proposed cross-modal fusion expert incorporates the dynamic routing weights of expert subnetworks from FVLMoE module of Forcevla \cite{yu2025forcevla}. Each expert subnetwork is specialized for a distinct phase of task execution and is adaptively activated according to high-level task instructions and low-level interaction feedback. In addition, we introduce a force-history sequence encoder and a multi-stage training scheme to further enhance performance. Combined with adversarial shared teleoperation, the proposed framework provides a system-level pipeline spanning data collection and end-to-end policy learning and control in contact-rich tasks.

\begin{figure*}[!t]
  \centering
  \includegraphics[width=380pt]{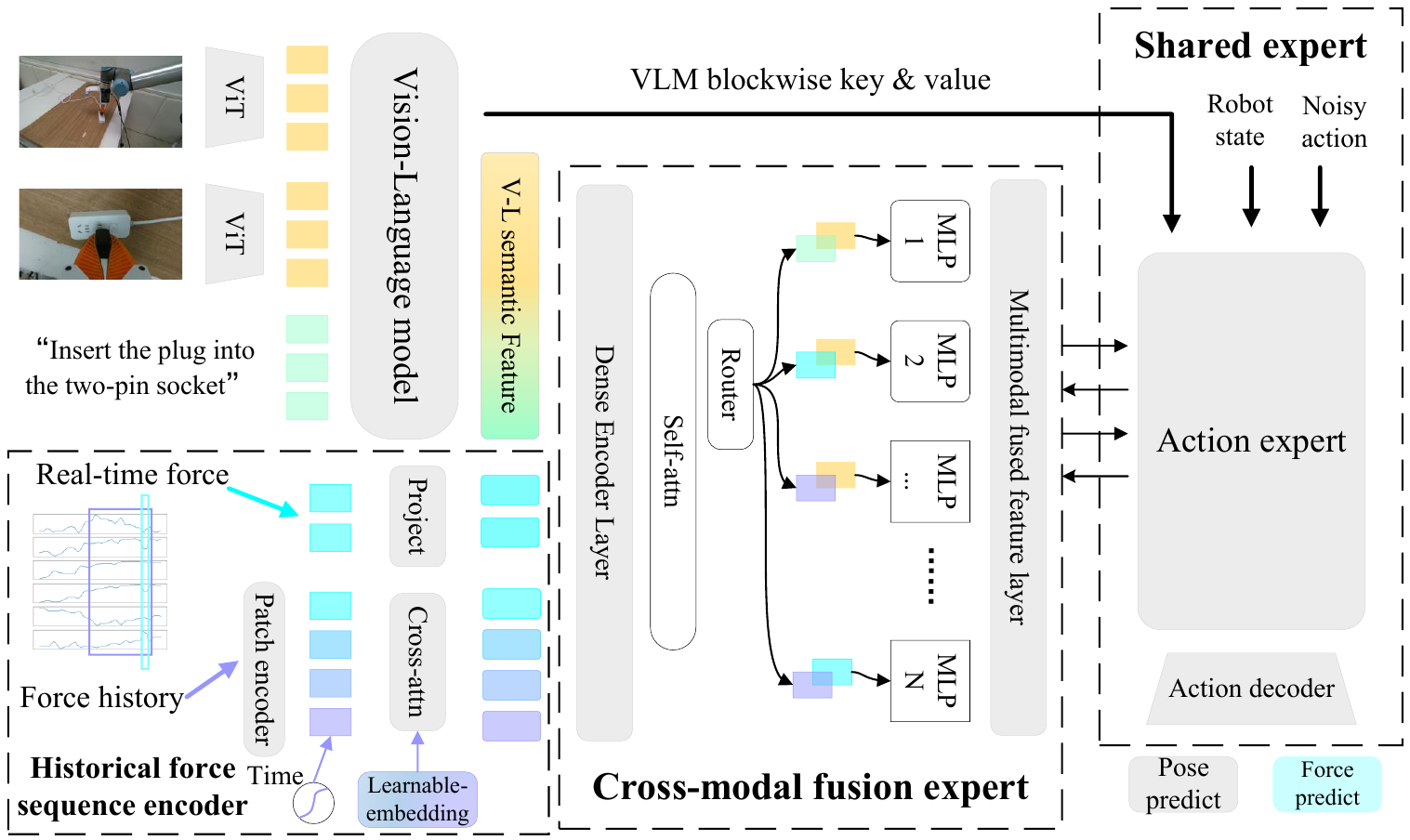}
  \caption{Overview of our control-aware VLA model. Built upon the ${\pi _0}$ pretrained model, historical force sequence encoder and cross-modal fusion expert are proposed to ensure efficient modality fusion and high-quality action generation via flow based shared expert.}
  \label{fig:architecture}
\end{figure*}

\subsection{Problem formulation}

We formulate the robotic manipulation as learning a control-aware VLA policy $\pi ({A_t}|{O_t},L)$ conditioned on a language instruction $L$. At timestep $t$, the observation ${O_t} = {\left\{ {I_t^b,I_t^h,{s_t},{F_t}} \right\}_t}$ consists of multi-view RGB images, proprioceptive state ${s_t} = \left\{ {{p_t},{g_t},{f_t}} \right\} \in {\mathbb{R}^{13}}$ (TCP pose, gripper width and real-time end-effector force), and historical force ${F_t} = \left\{ {{f_t}, \ldots ,{f_{t - h + 1}}} \right\} \in {\mathbb{R}^{h \times 6}}$ (for a time horizon $h$). Specifically, both the TCP pose within $p_t$ and the end-effector force $f_t$ are represented by 3D Cartesian coordinates and three Euler angles. A velocity field ${v_\theta }$ is predicted to iteratively denoise a Gaussian sample into a valid action sequence ${A_t} = \left\{ {{a_t},{a_{t + 1}}, \ldots ,{a_{t + H - 1}}} \right\}$ (for a time horizon $H$) consisting of predicted action ${a_t} = \left\{ {{p_t},{g_t},{f_t}} \right\}$, minimizing the objective:

\begin{equation}
{\mathcal{L}_{action}}(\theta ) = {\mathbb{E}_{\tau ,{O_t},\varepsilon }}{\left\| {{v_\theta }(A_t^\tau ,{O_t}) - ({A_t} - \varepsilon )} \right\|^2}
\end{equation} which sample Gaussian noise $\varepsilon \sim \mathcal{N}(0,I)$ and a time step $\tau  \in [0,1]$, forming a flow $A_t^\tau  = \tau {A_t} + (1 - \tau )\varepsilon $.

For compliant interaction, given the VLA-predicted $x_d$ as the desired trajectory, the robot's end-effector is controlled to follow the actual trajectory $x$ in Cartesian space:
\begin{equation}
{M_d}\ddot x + {D_d}\dot x + {K_d}\left( {x - {x_d}} \right) = {F_{ext}}
\end{equation}where ${K_d} \in {{\mathbb{R}}^{3 \times 3}}$ and ${D_d} \in {{\mathbb{R}}^{3 \times 3}}$ are the stiffness and damping factors. ${M_d} \in {{\mathbb{R}}^{3 \times 3}}$ represents the inertia factors, which is often ignored. The ideal tracking performance can lead to $\lim_{t \to +\infty} x(t) \to x_r(t)$, assuming that the actual trajectory tracks the reference trajectory $x_r$ effectively by the servo controller \cite{zeng2025robot}. The optimal interaction behavior $x_r$ of the robot with the real-time force is our control objective.

\subsection{Shared Teleoperation Data Acquisition}

A critical limitation in teleoperating position-controlled robots with non-haptic input devices, such as gamepads or 3D mice, is the sensitivity to positional errors during contact-rich manipulation. Small deviations relative to a rigid environment can trigger excessive interaction forces, frequently causing the robot to enter a protective stop state. Thus, this section proposes a shared teleoperation method based on compensated virtual impedance, which is equivalent to reduce the environmental impedance shown in Fig.3. Intuitively, for the same interaction force, a softer environment allows a larger contact displacement between the robot and the environment.

\begin{figure}[h]
  \centering
  \includegraphics[width=280pt]{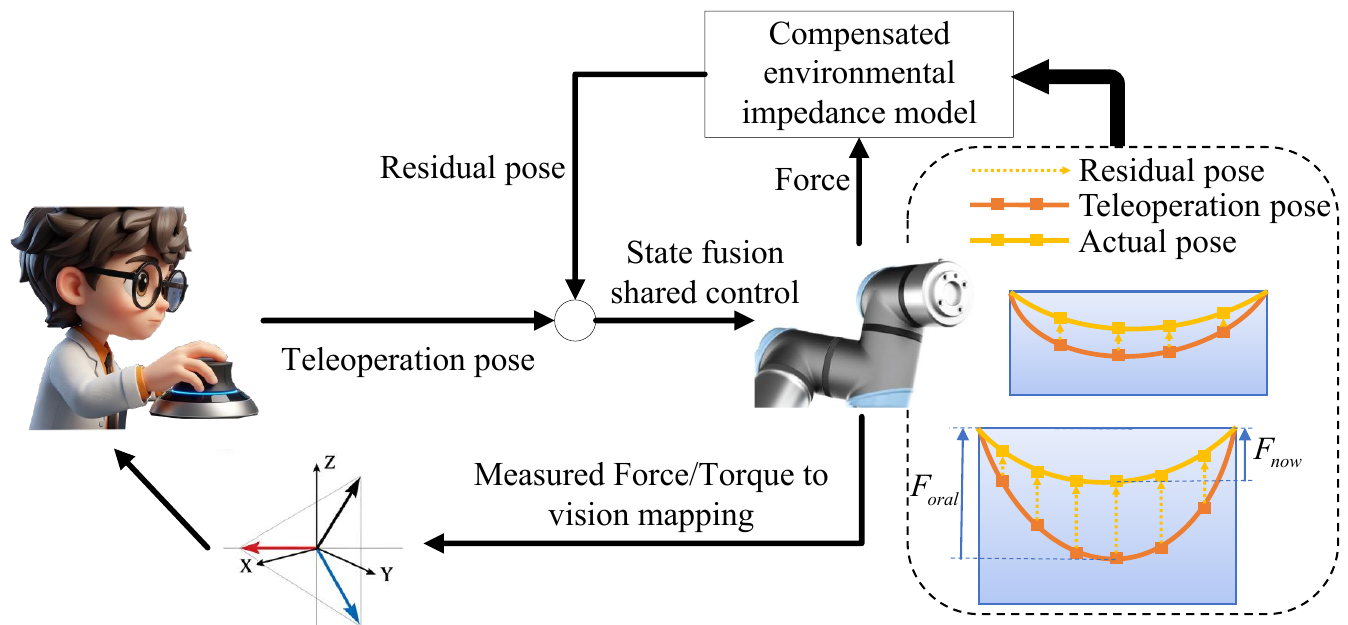}
  \caption{The scheme of data collection based on shared teleoperation method.}
  \label{fig:architecture}
\end{figure}

Firstly, when ignoring the inertia term $M_e \ddot{x}$ and the damping term $D_e \dot{x}$, the environmental impedance equation is given by:
\begin{equation}
    -F_e = K_e (x - x_e) = K_e \Delta x_o
\end{equation}
where $x = x_{tele}$ represents the actual position of the teleoperated robot. For the current displacement $\Delta x_o$ and the desired displacement $\Delta x_d$, we introduce a compensated virtual impedance $F_e = K_{adm} (\Delta x_d - \Delta x_o)$. Substituting this into the environmental dynamics equation, we obtain:
\begin{equation}
    -F_e = K_e \Delta x_d = K_e \left( K_{adm}^{-1} F_e + \Delta x_o \right)
\end{equation}
Rearranging the terms yields:
\begin{equation}
    -\left( I + K_e K_{adm}^{-1} \right) F_e = K_e \Delta x_o \implies -F_e = K_{enew} \Delta x_o
\end{equation}
Since $K_{e}$ and $K_{adm}$ are diagonal positive-definite matrices, the spectral radius of $\left( I + K_e K_{adm}^{-1} \right)^{-1}$ is less than 1. Therefore, we have:
\begin{equation}
    K_{enew} = K_e \left( I + K_e K_{adm}^{-1} \right)^{-1} < K_e
\end{equation}
After introducing the compensated virtual impedance $K_{adm}$ and actively utilizing force feedback to adjust the reference pose, the teleoperation system exhibits an equivalent stiffness that is softer than the actual environment. And the robot’s actual proprioceptive state during demonstration is recorded.

\subsection{Historical Force Sequence Encoder}

The historical force sequence accumulated over the inference observation interval encapsulates significant contact-state information (e.g., contact loss, force derivatives, peak magnitudes, and temporal trends), which is critical for determining subsequent actions and feedforward forces in contact-rich tasks. Thus, this section introduces a dedicated module designed to summarize historical force dynamics, supplementing real-time force observations. ${f_t} \in {\mathbb{R}^6}$ denotes the measured F/T at time step $t$. Given a history window of length $h$, directly inputting the raw sequence into the policy proves inefficient and sensitive to scaling. Furthermore, due to the absence of explicit supervision, it may fail to implicitly extract the requisite target features from raw data.

To address this, inspired by Causal Observation Perceiver \cite{xia2025cage}, we partition ${F_t}$ into $\frac{h}{p}$ contiguous segments of length $P$. Each segment is mapped to a latent representation by a shared force-patch encoder (implemented as a MLP), which captures local temporal patterns and correlation between force and torque channels. We then add temporal positional encodings to these segment embeddings to explicitly encode time ordering. Inspired by token-based summarization mechanisms, we introduce a learnable force token that performs cross-attention over all segment embeddings. The resulting attended token serves as a compact descriptor of the entire historical force sequence, aggregating long-horizon interaction cues, contact transitions, and magnitude trends into a fixed-dimensional feature. This historical force token is subsequently concatenated with visual and/or state tokens and processed by the downstream backbone, enabling the policy to exploit rich time and force sequence for decision making without incurring prohibitive computational overhead from the full force history.

\subsection{MoE-Based Feature Fusion with Multi-Stage Learning Strategy}

To effectively integrate high-frequency force feedback without compromising the pretrained semantic capabilities of the VLA backbone, this section proposes a multi-stage learning strategy for the model. Unlike prior approaches that train multimodal components simultaneously from scratch \cite{yu2025forcevla}, we decouple the learning process into two distinct phases. In the first stage, we finetune the $\pi_0$ \cite{black2410pi0} base model using block-wise attention masks, focusing on aligning vision, language, and proprioceptive action spaces. This establishes a robust policy capable of general manipulation based on visual affordances.

In the second stage, we introduce the cross-modal fusion expert as a residual branch to the action expert, which fuses F/T data with visual-linguistic embeddings via a sparse MoE mechanism. By optimizing the shared expert jointly using the loss function in (1) in this stage, the model learns to modulate the established action trajectories with fine-grained contact dynamics. It ensures that force-aware adjustments are learned as a refinement to the visual policy, preventing modality competition and enhancing stability in contact-rich tasks.

\begin{figure*}[t]
  \centering
  \includegraphics[width=420pt]{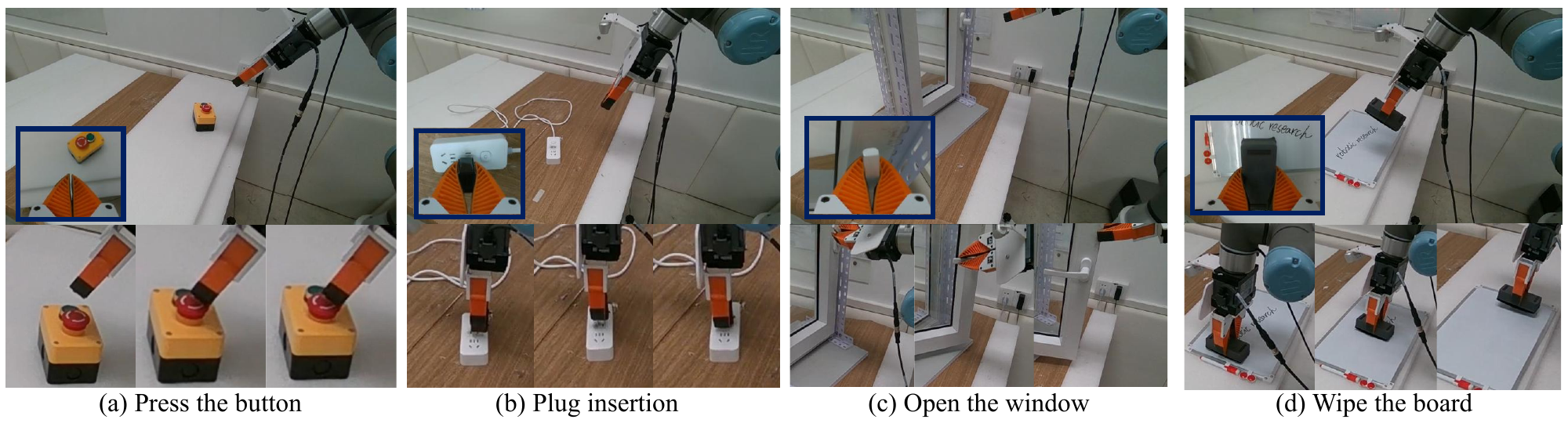}
  \caption{Overview of task setups used in evaluation.}
  \label{fig:architecture}
\end{figure*}

\subsection{Adaptive Compliance Controller}

\begin{figure}[h]
  \centering
  \includegraphics[width=280pt]{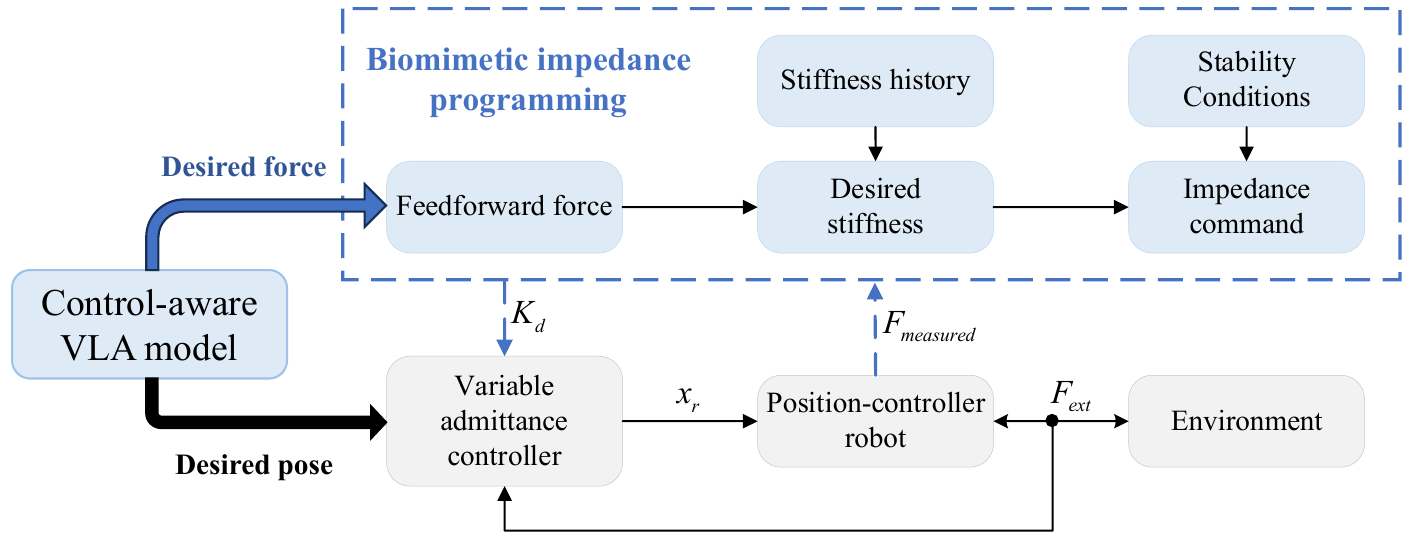}
  \caption{Diagram of the adaptive compliance controller, which receives desired action sequences inferred by the control-aware VLA model as input.}
  \label{fig:architecture}
\end{figure}

Following the acquisition of an action chunk from the policy rollout, an asynchronous action trajectory layer interpolates the discrete trajectory into a continuous sequence of desired action commands at a higher frequency as shown in Fig.5. Specifically, orientations are interpolated on the Lie group SO(3) employing spherical linear interpolation, which follows the geodesic path between rotational states.

Considering that human motor control commands comprise both feedback and feedforward components \cite{yang2011human}, we build upon the adaptive compliance controller \cite{zeng2025robot} to execute the desired actions denoted in (2), which is defined as:

\begin{equation}
{F_{ext}} = \underbrace {{K_d}(x - {x_d}) + {D_d}\dot x}_{{\text{Feedback }}} + \underbrace {{F_d}}_{{\text{Feedforward }}}
\end{equation}where ${K_d}$ and ${D_d}$ are typically passive because environmental impedance is variable and difficult to be measured. To maintain compliant interaction and control stability, QP based passive impedance control is often employed for desired force tracking \cite{zhao2022hybrid,beber2024passive}. Mathematically, another QP formulation based on force tracking error and stiffness variation rate \cite{zeng2024hierarchical} can be interpreted as applying a convex QP solver to compute a constrained gradient update law as follows:

\begin{equation}
{K^t} = {K^{t - 1}} - \eta \nabla J\left( {{K^{t - 1}}} \right)
\end{equation}where $J$ denotes the force tracking cost function and $\eta  > 0$ represents the update step size. However, considering the potential for stochastic fluctuations in the desired actions and forces generated by the CC-VLA action expert, we opted against using QP to solve for a target stiffness for precise force tracking. Instead, inspired by the principles of Resilient Propagation, we formulated a heuristic adaptation rule designed to minimize the force tracking error as follows:

\begin{equation}
{K^t} = {K^{t - 1}} - \alpha  \odot \operatorname{sign} \left( {\left| {F_{{\text{err}}}^k} \right| - \left| {F_{{\text{err}}}^{k - 1}} \right|} \right),{K^0} = {K_{\min }}
\end{equation}where $F_{{\text{err}}}^t = F_e^t - F_d^t$ denotes the tracking force error, $\alpha $ represents the step size in each direction, $ \odot $ indicates element-wise multiplication, and $\operatorname{sign} ( \cdot )$ denotes the element-wise sign operation. The step size is defined as follows:

\begin{equation}
\alpha  = \lambda  \cdot \left| {F_{err}^k} \right|
\end{equation}where $\lambda $ is a control parameter. Then a low-pass filter and output saturation are applied to the resulting stiffness ${K_d}$:

\begin{equation}
\left\{ \begin{gathered}
  {K^t} = \beta K_{{\text{o}}ri}^t + (1 - \beta )K_{{\text{o}}ri}^{t - 1},{\text{ }}0 < \beta  < 1 \hfill \\
  {K_d} = {\Pi _{\left[ {{K_{\min }},{\text{ }}{K_{\max }}} \right]}}\left( {{K^t}} \right) \hfill \\ 
\end{gathered}  \right.
\end{equation}where ${\Pi _{[a,b]}}( \cdot )$ denotes the element-wise projection onto the interval $\left[ {a,b} \right]$. Inspired by \cite{kronander2016stability}, the damping matrix ${D_d} = 2\zeta \sqrt {{K_d}}$ is selected to satisfy the critical damping condition with a fixed natural index $\zeta $ to guarantee the fastest convergence of the error without oscillations. To ensure global asymptotic stability of the time-varying stiffness controller, the stiffness variation is constrained by output limiting such that ${K_d} + \omega {\dot D_d} - {\omega ^2}{K_d}$ remains negative semidefinite. \cite{zeng2024hierarchical}

\begin{equation}
\dot K < \frac{{2\omega {{\sqrt K }^3}}}{{\sqrt K  + 2\zeta \omega }}
\end{equation}where $\omega $ is a positive vector, so as to satisfy the Lyapunov function with position–velocity cross terms constructed in \cite{kronander2016stability}.

\section{Experiments}
We evaluated the proposed CC-VLA across four real-world tasks, comparing its performance against the prevailing methods involving Diffusion Policy \cite{chi2025diffusion}, ${\pi _0}$ \cite{black2410pi0}, ${\pi _0}$ wi F(with force signals directly concatenated to state inputs), ${\pi _{05}}$ \cite{intelligence2025pi_} and ForceVLA \cite{yu2025forcevla}. Additionally, we conducted an ablation study to assess the contributions of the control-aware VLA model and the adaptive compliance controller.

\begin{figure*}[t]
  \centering
  \includegraphics[width=390pt]{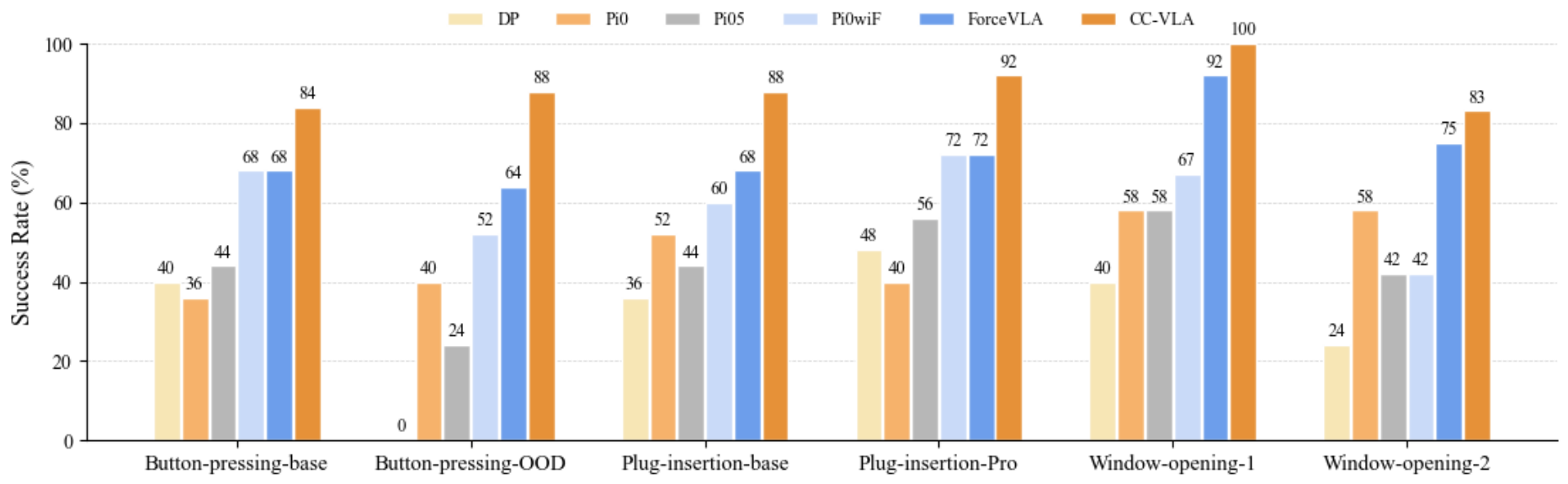}
  \caption{Success rates across different methods for three contact-rich manipulation tasks.}
  \label{fig:architecture}
\end{figure*}

\begin{figure*}[t]
  \centering
  \includegraphics[width=450pt]{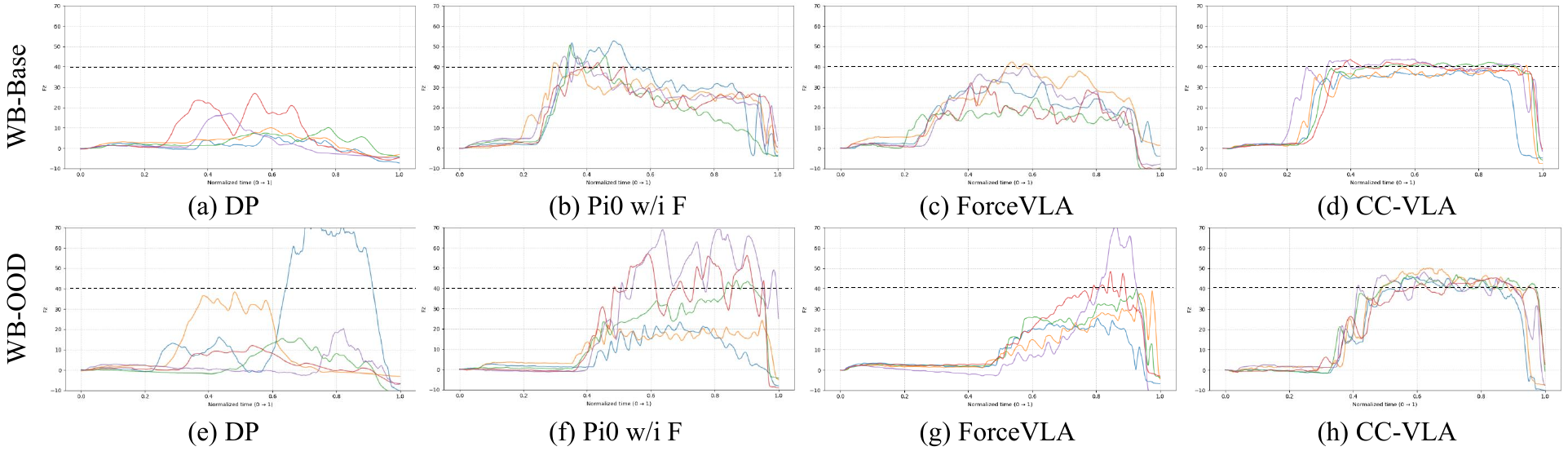}
  \caption{Z-axis force profiles recorded during the first five trials of wipe the board task using various methods. Notably, trajectory execution times have been normalized to account for discrepancies in the initial robot pose and board placement. The dashed line indicates the demonstrated reference force of 40 N. Data for methods Pi0 and Pi05 are omitted, as these models triggered the robot's protective stop mechanism due to excessive interaction forces. }
  \label{fig:architecture}
\end{figure*}

\subsection{Experimental Settings}
1) Experimental Setup: 
We employed a position-controlled robotic arm (UR5e) equipped with a wrist-mounted camera (RealSense D435), a UMI-like gripper, and a 6-DoF force/torque sensor. Another camera was mounted on the side of the robotic arm to provide a frontal view of the workspace. The UR5e robot was controlled via the UR RTDE library running on a host computer equipped with an Intel i7-13700KF CPU. The VLA model, with a sparse MoE structure of 4 experts, performed inference once per second with an action horizon of 40, while the force sequence was sampled every 0.1 s. The control frequency was 500 Hz, and the stiffness bounds ranged from 400 N/m to 2500 N/m.

2) Task definition: 
To evaluate the efficacy of CC-VLA across diverse contact-rich manipulation scenarios, we selected four representative tasks: pressing an emergency stop button, inserting a charging plug, opening a rotary window, and wiping a whiteboard with constant force (see Fig. 4). Recognizing that contact-rich tasks generally necessitate either force perception or force tracking capabilities, our evaluation explicitly analyzes force tracking performance in the board-wiping task: a critical metric often overlooked in similar studies. Each task addresses specific challenges:

\textbf{Button-pressing (PB)}: It requires force feedback, as multi-view images and proprioception are insufficient to determine the button's engagement state, especially in OOD settings.

\textbf{Plug-Insertion (PI)}: This multiple peg-in-hole task involves exploratory insertion that requires delicate force perception to minimize contact impact, followed by precise visual alignment and sufficient interaction force to ensure complete insertion.

\textbf{Window-opening (OW)}: It is a long-horizon, high-precision, contact-rich manipulation task. Without compliant motion adaptation, excessive normal forces or torques can easily trigger the robot’s protective emergency stop (self-lock).

\textbf{Board-wiping (WB)}: Representative of sanding and polishing applications, this task requires the robot to manipulate an eraser while maintaining a relatively constant normal force against the surface.

3) Experimental groups: 
To train the button‑pressing, window‑opening, and board‑wiping tasks, we collected 50 expert demonstrations per task. For the plug‑insertion task, we expanded the dataset to 100 demonstrations to improve the overall success rate. It was observed that this fine‑grained peg‑in‑hole task required more demonstrations to account for a wider range of positional variations during testing.

We first conducted permutation testing experiments within the positional distribution range of the demonstration data for four tasks (denoted as PB-base, PI-base, OW-base, and WB-base). Furthermore, we established out-of-distribution height generalization experiments for button-pressing and board-wiping (PB-OOD and WB-OOD). In these scenarios, the placement heights of the button and board were set 6 cm lower than those used during data collection. This requires the model to overcome height generalization challenges through force perception and motion control, thereby testing its robustness. For the plug-insertion task, we introduced a proximal start experiment (PI-pro), wherein the plug is initialized near the socket but in a position where the wrist-mounted camera cannot capture the socket holes. This mimics the force-sensing exploratory behavior exhibited by humans when locating an out-of-sight socket. Finally, as board-wiping is a force-tracking control task, we analyzed the interaction force variations throughout the entire process to verify the superior force-tracking control performance of the proposed CC-VLA. Each test condition comprised 24--25 trials.

\subsection{Results}

\begin{figure*}[t]
  \centering
  \includegraphics[width=450pt]{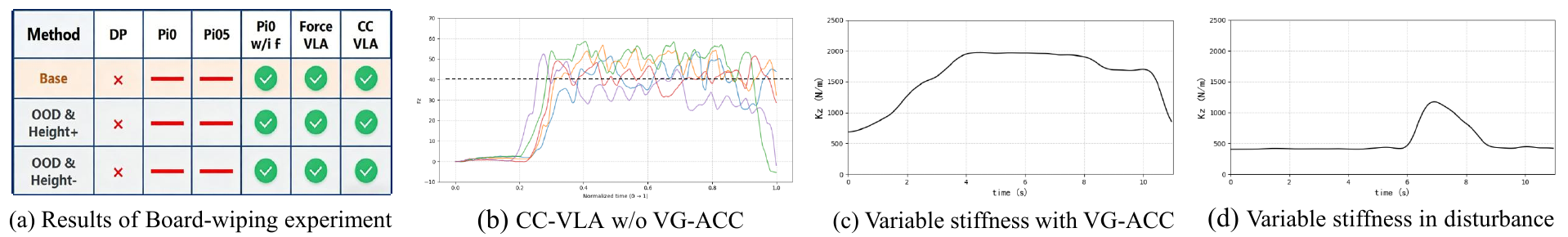}
  \caption{(a) Task outcomes: The red horizontal line indicates a protective stop triggered by excessive interaction forces; the cross mark $ \times $ denotes incomplete wiping; and the check mark $\surd $ signifies successful task completion. (b) Z-axis force profiles observed when executing pose sequences generated solely by control-aware VLA inference. (c)–(d) The corresponding stiffness profiles. VG-ACC denotes the proposed VLA-guided adaptive compliance controller.}
  \label{fig:architecture}
\end{figure*}

As demonstrated in Fig. 6, CC-VLA achieves an average success rate of 89.2\% across all six conditions, significantly outperforming all baseline configurations (DP: 31.3\%, Pi0: 47.3\%, Pi05: 44.7\%, Pi0 wi F: 60.2\%, ForceVLA: 73.2\%). Specifically, Window-opening-1 refers to the phase in which the robot grasps the handle and pulls the window open to a specified angle. Window-opening-2 then corresponds to the phase where the robot braces its gripper against the window frame edge to open the window fully. These results underscore the significant advantage of effectively integrating multimodal VLA models, which is capable of sequential force perception with reactive force control.

In the WB-base and WB-OOD experiments, conducted within the shared teleoperation method detailed in Section III.B and force-data preprocessing pipeline, a constant normal force of 40 N was maintained during demonstrations to ensure effective residue removal. Fig. 8(a) reports the completion status of each comparative method. Additionally, the Z-axis interaction forces were monitored, with the trajectories of the first five trials plotted in Fig. 7. The observations reveal that the CC-VLA is the only method capable of robustly tracking the 40 N desired force. Correspondingly, the tracking errors $\frac{{\sum\nolimits_0^n {\left| {{F_{ext}} - {F_{demo}}} \right|/\left| {{F_{demo}}} \right| \times 100\% } }}{n}$ between the actual Z-axis force ${F_{ext}}$ and the desired force ${F_{demo}}$ are quantified in Table I. When the instantaneous interaction force exceeds 75 N, the robot triggers a protective stop, as observed in the Pi0 and Pi05 experimental groups. In contrast, the DP condition fails to achieve satisfactory wiping performance because the interaction force is too low or results in intermittent contact loss. Furthermore, Table II presents an ablation study evaluating the impact of the proposed variable impedance law and the VLA-guided adaptive compliance controller.

\begin{table}[h]
    \centering
    \caption{Performance (\%) of demonstrated force tracking error across different methods.}
    \label{tab:base_ood_results}

    \resizebox{0.85\columnwidth}{!}{%
    \begin{tabular}{lcccccc}
        \toprule
        & DP & Pi0 & Pi05 & Pi0 wi force & ForceVLA & CC-VLA \\
        \midrule
        WB-Base & 83.83\% & NA & 76.12\% & 28.10\% & 35.57\% & 5.52\% \\
        WB-OOD  & 79.33\% & 81.20\% & 82.19\% & 38.15\% & 52.33\% & 8.78\% \\
        \bottomrule
    \end{tabular}%
    }

\end{table}

\begin{table}[h]
\centering
\caption{Performance (\%) of demonstrated force tracking error for ablation experiments.}
\label{tab:base_controllers}

\resizebox{0.5\columnwidth}{!}{%
\begin{tabular}{lcccc}
\toprule
& wo VG-ACC & wi $K_{\mathrm{low}}$ & wi $K_{\mathrm{high}}$ & wi $K_{\mathrm{vari}}$ \\
\midrule
Base & 17.73\% & 7.92\% & 13.30\% & 5.52\% \\
\bottomrule
\end{tabular}
}

\end{table}

\subsection{Discussion}

1) Comparison study:
The results of the PB-Base, PB-OOD, PI-Base, and PI-Pro experiments demonstrate the effectiveness of the proposed CC-VLA for force-sensing tasks and its robustness under OOD conditions. The OW-Base task further highlights the high performance of CC-VLA in compliant tasks that require precise long-range manipulation. Moreover, the WB-base and WB-OOD experiments show that integrating the control-aware VLA with the adaptive compliance controller enables stable force tracking in OOD scenarios, representing an experimental setting and capability that have not been explicitly evaluated in prior work.

2) Ablation study:
To evaluate the proposed multi-stage training method, we conducted ablation studies on the PB-Base and PI-Base tasks, using CC-VLA without the multi-stage training method as the baseline. For the PB-Base task, the proposed training method improved the success rate by 12\%. In particular, it reduced the false-pressing pose rate from 28\% to 8\%, suggesting that the multi-stage training paradigm better preserves the model's spatial awareness. In the PI-Base scenario, our method attained a 76\% success rate, representing an 8\% improvement in success rate. For the historical force-sequence encoder module, a PI-Pro task ablation study shows that the success rate increases from 72\% to 92\%, demonstrating that this module enables the policy to exploit fine-grained force sensing and manipulation-state estimation to guide action execution. Additionally, results from the board-wiping ablation study demonstrate the practical value of the proposed compliance controller and variable stiffness mechanisms.

3) Safety-aware study:
Experiments confirm that by learning human action-adjustment skill from demonstrations under excessive contact forces, the proposed model effectively balances safe interaction forces with task success rates. Furthermore, the variable impedance law ensures compliance in the presence of excessive forces (Figs. 8(b) and 7(d)) and external disturbances (Fig. 8(d)). From a system-level perspective, the desired force generated by the flow-based VLA follows a distribution similar to that of the demonstration forces under the same visual conditions. Even when the VLA infers a desired pose that could lead to hazardous contact due to perception or state-estimation errors, the proposed controller limits excessive contact forces, thereby ensuring compliant interaction and enhancing the safety of the hierarchical system.

\section{Conclusion}
In this work, we address the issues of force-action adjustment latency and poor force-tracking accuracy observed in existing VLA models by proposing a control-aware VLA framework with compliance controller designed to achieve reactive force–position control. To enhance both action and force prediction in VLA models, we introduce a two-stage training strategy and a force history sequence encoder, together with a multimodal MoE module, improve the perception of force sequences and strengthen the VLA model’s spatial perception capability. Experimental results demonstrate that the proposed CC-VLA significantly outperforms existing methods on contact-rich tasks under various challenging conditions, as well as in terms of force-tracking performance. Moreover, the proposed approach can be readily applied to the fine-tuning of other pretrained VLA models, exhibiting good scalability. Nevertheless, the current method relies heavily on precise force control and temporal consistency during the acquisition of teleoperation demonstration data. Furthermore, it focuses exclusively on object-level variable impedance manipulation. Future work will investigate the integration of finger-level gripper control with object-level end-effector motion.

% \bibliography{reference}
% \input{preview_accent.bbl}

% \bibliographystyle{unsrtnat}
\bibliographystyle{IEEEtran}
\bibliography{reference}
\end{document}